\documentclass[10pt,twocolumn,letterpaper]{article}

\usepackage{iccv}
\usepackage{times}
\usepackage{epsfig}
\usepackage{graphicx}
\usepackage{amsmath}
\usepackage{amssymb}

\usepackage{multirow}
\usepackage[table,xcdraw]{xcolor}
\usepackage{float}
\definecolor{bad}{rgb}{0.25098039215686274, 0.0, 0.29411764705882354}
\definecolor{good}{rgb}{0.0, 0.26666666666666666, 0.10588235294117647}

\usepackage[breaklinks=true,bookmarks=false]{hyperref}

\iccvfinalcopy % *** Uncomment this line for the final submission

\def\iccvPaperID{****} % *** Enter the ICCV Paper ID here

\ificcvfinal\fi

\begin{document}

%%%%%%%%% TITLE
\title{Simultaneous Segmentation and Recognition: Towards more accurate Ego Gesture Recognition}

\author{Tejo Chalasani\\
V-Sense\\
Trinity College Dublin\\
{\tt\small chalasat@tcd.ie}
\and
Aljosa Smolic\\
V-Sense\\
Trinity College Dublin\\
{\tt\small smolica@scss.tcd.ie}
}

\maketitle
\ificcvfinal\thispagestyle{empty}\fi

%%%%%%%%% ABSTRACT
\begin{abstract}
   Ego hand gestures can be used as an interface in AR and VR environments. While the context of an image is important for tasks like scene understanding, object recognition, image caption generation and activity recognition, it plays a minimal role in ego hand gesture recognition. An ego hand gesture used for AR and VR environments conveys the same information regardless of the background. With this idea in mind, we present our work on ego hand gesture recognition that produces embeddings from RBG images with ego hands, which are simultaneously used for ego hand segmentation and ego gesture recognition. To this extent, we achieved better recognition accuracy (\(96.9\%\)) compared to the state of the art (\(92.2\%\)) on the biggest ego hand gesture dataset available publicly. We present a gesture recognition deep neural network which recognises ego hand gestures from videos (videos containing a single gesture) by generating and recognising embeddings of ego hands from image sequences of varying length. We introduce the concept of simultaneous segmentation and recognition applied to ego hand gestures, present the network architecture, the training procedure and the results compared to the state of the art on the EgoGesture dataset \cite{Zhang2018}.
\end{abstract}

\section{Introduction}
\label{sec:1_intro}
Head-mounted AR and VR devices like Magic Leap One, Microsoft HoloLens, Oculus Rift and Samsung Gear VR are becoming widely available. While the usage of joysticks is certainly one way to interact with virtual objects in such devices, hand gestures would facilitate more intuitive and natural interactions with virtual objects in AR and VR devices. Thus, recognising hand gestures as seen by the cameras on AR/VR, known as ego hand gestures becomes a problem worth exploring.

Ego gesture recognition in addition to traditional gesture recognition \cite{Ramamoorthy2003, Neverova2014, Molchanov2016} brings in more challenges like ego head motion, obstructed or partial view of hands in the camera. The existing algorithms which work for normal gesture recognition \cite{Suk2010, Nagi2011, Tang2013, Neverova2014} can not deal with the above challenges that ego gesture recognition poses \cite{Serra2013}.

Ego gesture recognition prior to deep learning used handcrafted features calculated from spatially segmented hands for static ego gesture recognition \cite{Serra2013}. Feature descriptors like Histogram of Gradients (HoG) along with motion descriptors like Histogram of Flow (HoF) and optical flow with dense feature trajectories and homography compensation for head motion were used to recognise dynamic ego hand gestures \cite{Baraldi2014, Hegde2016}. However, in all the approaches above the number of gestures recognised were very small (up to 10). With the advent of deep learning there has been some research in applying different network architectures for recognising ego gestures \cite{Cao2017, Jain2017, tejo2018}.

Cao \etal~ \cite{Cao2017} introduced an EgoGesture dataset with 83 gestures performed by a large number of subjects and a network architecture with 3DConvolutional Neural Networks (3DCNNs), Spatio-Temporal Transformer Modules (STTM) and Long/Short Term Memory(LSTM) to get the state of the art classification results on the dataset. The 3DCNNs conceptually calculates the motion features and STTM compensates for the head motion. In our architecture, we forgo the specific use of 3DCNNs, STTM and introduce embeddings for ego hands that can be simultaneously used for ego hand segmentation and ego gesture recognition. The primary idea is to encode RGB images with ego hands and find embeddings that correspond only to ego hands. Then use these embeddings generated from a sequence of images in a recurrent neural network to recognise the corresponding ego hand gesture. We elaborate on the idea of simultaneous segmentation and recognition in Section \ref{sec:3_decon} and provide the full network architecture in Section \ref{sec:4_network}. Once trained, our network only needs RGB images during inference and also has the ability to use any number of images in a sequence. The details about the experiments and results are provided in Section \ref{sec:5_exp}. 

\begin{figure*}
\begin{center}
\includegraphics[scale=0.8]{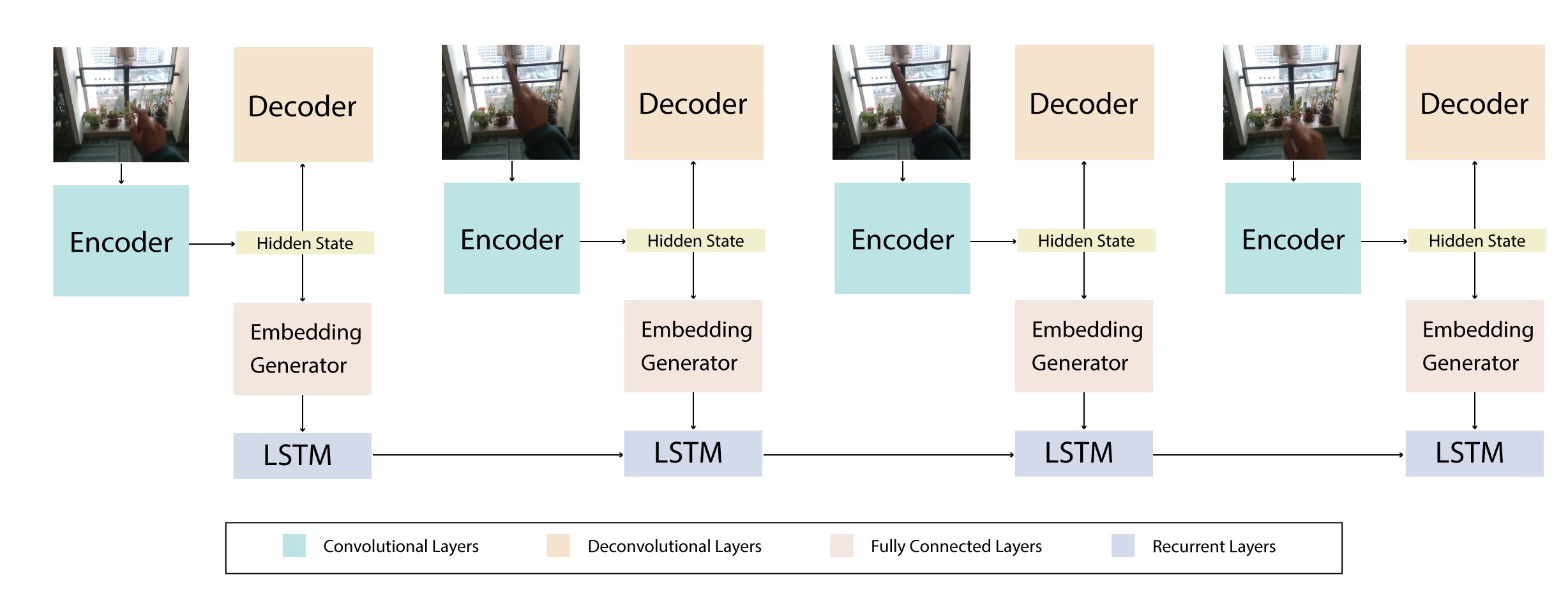}
\end{center}
   \caption{Our network architecture. The Decoder outputs a segmented ego hand map per image in the sequence. This part of the architecture is not needed once the network is trained. The sequence of images converted to embeddings corresponding to ego hand segmentation is used by LSTMs to recognise the ego gesture.}
\label{fig:3_network}
\end{figure*}

\textbf{Our contribution.} We propose the concept of simultaneous segmentation and recognition for achieving better ego hand gesture recognition, with emphasis on using the segmentation for better recognition accuracy and not vice versa. A new deep learning network (Figure \ref{fig:3_network}) architecture that generates embeddings which can be used for ego hand segmentation and gesture recognition at the same time is introduced. Our network improved considerably on the state of the art results for EgoGesture \cite{Zhang2018, Cao2017}, the biggest ego gesture dataset available publicly. Unlike the state of the art \cite{Cao2017} we do not restrict the sequence length to 40, but use any length sequence given as input from the dataset. We need only RGB modality during inference time for our results, whereas the state of the art uses both RGB and depth modalities for achieving their best scores.

\section{Previous Work}
\label{sec:2_prev_work}
\begin{figure*}[htb!]
\begin{center}
\includegraphics[scale=0.95]{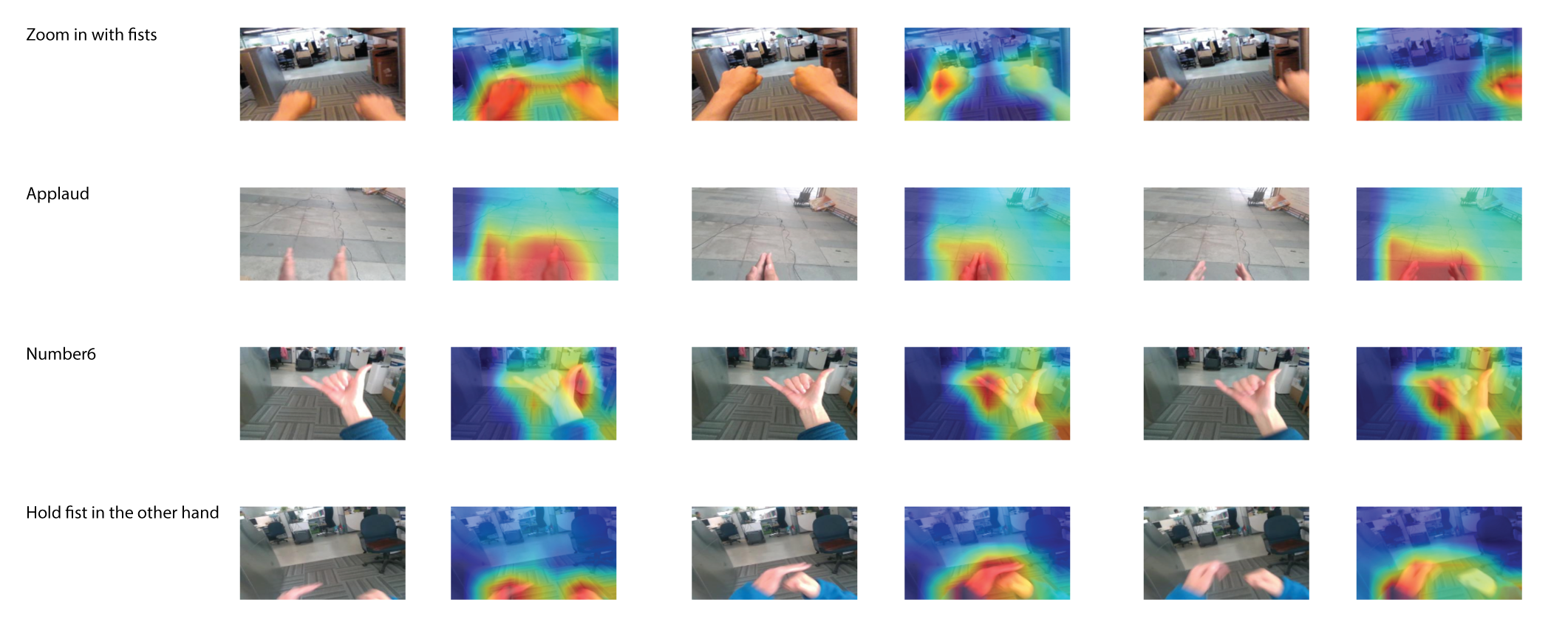}
\end{center}
   \caption{Gradient Class Activation Maps(Grad-CAM)\cite{SelvarajuDVCPB16} for showing the activations that correspond to the gesture. Each row has three RGB images and their corresponding Grad-CAM images from a sequence that belongs to a gesture. For illustration purposes we choose gestures that use both hands and only one hand, to show the robustness of the network. The class activation maps rightly highlights the areas that belong to ego hands, intuitively in some gestures as seen for Number6 gestures, the fingers on the hand contribute more towards identifying the gesture correctly.}
\label{fig:4_visualisation}
\end{figure*}

Traditionally hand gesture recognition used features calculated from segmenting hands. Algorithms like Simple Linear Iterative Clustering (SLIC) and colour histogram identification \cite{Suk2010, Serra2013} were used for segmenting hands. Once the hand is segmented, handcrafted spatial and spatio-temporal features like HoF and improved dense trajectories are calculated and used in Hidden Markov Models or Dynamic Bayesian Networks \cite{Moni2009, Wang2013}. To deal with the complexity of ego motion, Baraldi \etal ~\cite{Baraldi2014} suggested using homography compensation before calculating the handcrafted features. Though not directly related to gesture recognition, Chang \etal \cite{chang2016} has done similar work for detection of finger-writing from ego-centric videos. From segmented hands, they detect the finger tip and its trajectory, then extract spatio-temporal Hough forest features to assign a class label to the trajectory.

All deep learning methods require large amounts of annotated training data for networks to generalise well. An improvised ChaLearn dataset was introduced by Escalera \etal ~\cite{Escalera2014} to encourage deep learning research in action and gesture recognition. It has multimodal data containing depth and intensity for each hand separately and full body pose information all synchronised together. Neverova \etal \cite{Neverova2014} proposed a multi-scale architecture that could deal with varied gesture duration using all the modalities present in the ChaLearn dataset. More recently 3DCNNs and recurrent neural networks (RNN) were used for gesture recognition by Molchanov \etal \cite{Molchanov2016}, setting benchmark performance for a dataset they introduced and the state of the art for the ChaLearn dataset. RNNs were used in this work to deal with varied gesture duration instead of the multi-scale approach proposed in \cite{Neverova2014}.

Zimmerman and Brox \cite{Zimmermann2017} proposed a deep neural network to estimate the pose of a hand in 3D using just an RGB image. Jain \etal \cite{Jain2017} used the pose predictions from this network and trained LSTMs to recognise 4 ego hand gestures. Chalasani \etal \cite{tejo2018} introduced the concept of using green screen for data augmentation for ego hands. They proposed a network architecture which was tested on a dataset they introduced (containing 10 gestures) and another small dataset containing 4 gestures published by Jain \etal in \cite{Jain2017}. Their work focused on increasing the dataset size through augmentation using green screen extraction process to generate features for ego hands from a small number of training samples.

Bambach \etal \cite{Bambach2015} proposed using CNNs for segmenting hands from egocentric videos and using the segmented hand mask images as input to another CNN to recognise ego hand activity. Zhu \etal \cite{zhu2016} used a similar approach to detect bounding boxes around hands and then finding hand shape masks and heat maps of wrist and palm locations within the bounding box. These heat maps and hand masks are then used in a different classifier to classify ego hand activities. More recently the method proposed by Khan and Borji \cite{Urooj2018} used a fine-tuned version of RefineNet\cite{Liu2017} in conjunction with Conditional Random Fields to achieve pixel-level hand segmentation and used the segmentation masks later with AlexNet for ego hand activity detection. Li \etal \cite{li_dong_2018} proposed the concept of recurrent tubelets proposal and recognition.In this approach the current area related to hand is extracted based on its previous location recurrently, and features are calculated on this extracted area. These features are then fed into a separate network for recognising gestures. In all the above approaches \cite{Bambach2015, zhu2016wacv, Urooj2018, li_dong_2018}, features were calculated on the extracted ego hand masks and then provided as an input to a different recognition system. Instead in our approach, we calculate features which can be both used for ego hand mask generation and ego gesture recognition simultaneously and also giving our network architecture ability to train end-to-end which wasn't possible in earlier approaches. 

The EgoGesture dataset \cite{Cao2017, Zhang2018} was introduced specifically for training, bench-marking and evaluating deep neural networks for recognising ego hand gestures. Cao \etal \cite{Cao2017} proposed an end-to-end learnable network that used 3D CNNs in conjunction with STTM and LSTMs, achieving the state of the art classification accuracy on the EgoGesture dataset. Considering the successful application of 3D CNNs and RNNs to gesture recognition in \cite{Molchanov2016}, Cao \etal extended the network architecture to include STTMs and Recurrent STTMs(RSSTMs). Inspired by spatial transformer networks \cite{jaderberg2015spatial}, STTMs transform a 3D feature map to compensate for the ego-motion that is introduced by head movement. Our network with decontextualised embeddings can efficiently deal with ego-motion without the need for movement compensation.

\section{Simultaneous Segmentation and Recognition}
\label{sec:3_decon}
\begin{figure*}
\begin{center}
\includegraphics[scale=0.8]{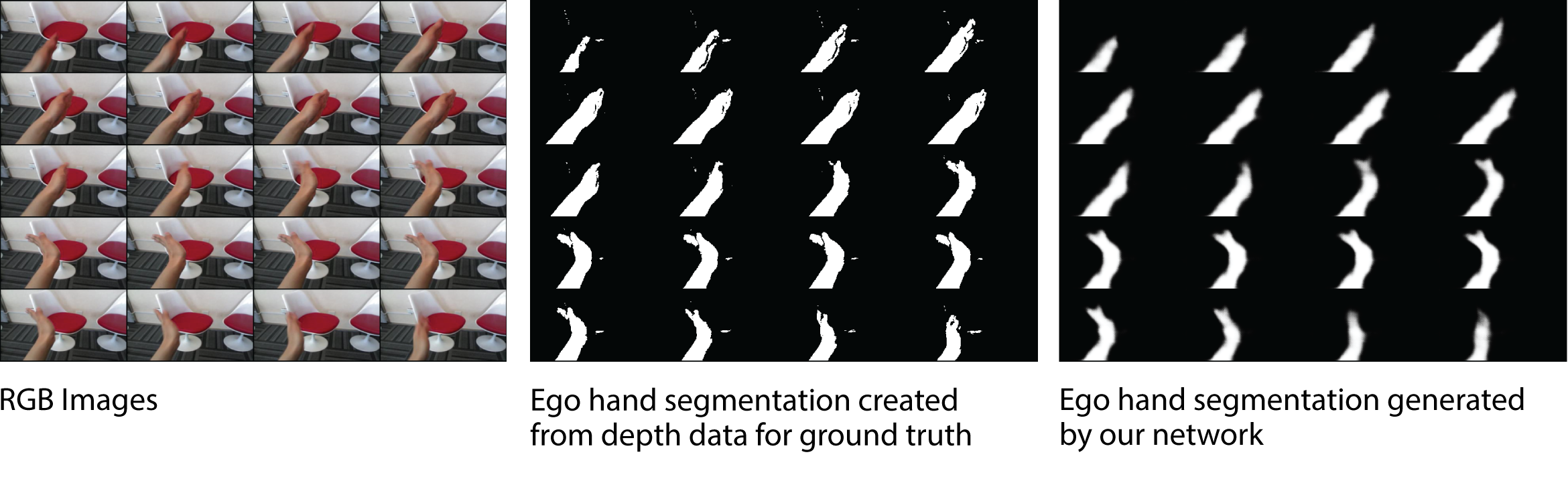}
\end{center}
   \caption{EgoGesture Dataset \cite{Zhang2018} does not provide explicitly annotated hand mask data. We threshold depth maps provided to extract pixels responsible for ego hands to generate ground truth data. This process can introduce noise, but our network learns to generate segmented ego hands without noise.}
\label{fig:2_Decon}
\end{figure*}

As discussed in Section\ref{sec:2_prev_work}, work done by Bambach \etal \cite{Bambach2015}, Zhu \etal \cite{zhu2016wacv}, Khan and Borji \cite{Urooj2018} has shown that segmented hand image can be a strong feature for recognising ego gestures. However, in all these approaches, features for recognition are calculated on the segmenting hands. To our knowledge, we are the first to propose using the same feature set for ego hand segmentation and gesture recognition simultaneously. We attempt to formalise this idea below.

Let $I_s$ be the sequence of images containing ego hands that we want to recognise an ego gesture from and $l$ its corresponding class label. The problem of ego gesture recognition can then be defined as finding a function $f$, such that it maps the given image sequence $I_s$ to $l$ as described in equation \ref{eq1:recognition}.

\begin{equation}
\label{eq1:recognition}
f(I_s) = l
\end{equation}

So far the approach to finding the function $f$ using hand segmentation has been (represented by equation \ref{eq2:current_approaches}), to find three functions $g, h, k$ where 
\begin{itemize}
    \item $g$ takes in a sequence of images $I_s$, produces a sequence of ego hand masks $M_s$.
    \item $h$ takes in the sequence of ego hand masks $M_s$ generated by $g$, extracts a set of features $X_s$, where $dim(X_s) \ll dim(I_s)$
    \item $k$ takes in the set of features $X_s$ generated by $h$, and maps it to the corresponding class label $l$
\end{itemize}
\begin{equation}
    \label{eq2:current_approaches}
    f(I_s) = g(I_s) \circ h(M_s) \circ k(X_s)
\end{equation}

We rephrase this problem as finding the feature set $E_s$ for the sequence of images $I_s$ which can be simultaneously used for finding their corresponding segmented hand masks $M_s$ and also the class label $l$. In our approach, we define 3 functions $a, b, c$, where
\begin{itemize}
    \item $a$ takes in a sequence of images $I_s$, produces a feature set $E_s$, where $dim(E_s) \ll dim(I_s)$
    \item $b$ and $c$ take in the feature set $E_s$, and simultaneously produce ego hand masks $M_s$ and class label $l$.
\end{itemize}
Equation \ref{eq3:our_approach} summarises the concept of simultaneous segmentation and recognition. One advantage we have over previous methods is that, once the functions $a, b, c$ are estimated, we do not need the function $b$ that generates ego hand masks for recognising the class label.

\begin{equation}
    \label{eq3:our_approach}
    f(I_s) = a(I_s) \circ c(E_s) \backepsilon b(E_s) = M_s
\end{equation}

We used deep neural networks to design our functions $a, b, c$. Autoencoders \cite{DeepLearning} are well studied and known for finding a reduced representation of an input, thus we used them for designing functions $a, b$ with a slight modification. In general the input and the output for an autoencoder are the same, however, we use an RGB image with ego hand as input and a segmented ego hand mask as an output. This strategy helps our network to find a reduced representation of ego hand masks from the input RGB images. 

There are two components to an autoencoder, an encoder network which encodes the input to a short intermediate feature vector and a decoder network which recovers the output from this intermediate representation. In our case function $a$ represents the encoder part of the network and function $b$ represents the decoder part of the network. We use the embedding generator and the LSTM as function $c$ which also takes in the intermediate feature vector as input and generates the class labels for sequences of images. We elaborate on the network architecture of each component in Section \ref{ssec:4_1_Architecture}. The end-to-end training for simultaneous segmentation and recognition is explained in Section \ref{ssec:4_2_Training}. 

Our network design and training approach yielded a considerable improvement compared to the state of the art on EgoGesture dataset \cite{Zhang2018}, the details of which are provided in Section \ref{sec:5_exp} along with ablation studies we performed.

\section{Network Architecture and Training Strategy}
\begin{table*}
\begin{center}
\begin{tabular}{|l|l|l|l|}
\hline
                             & \textbf{Input}      & \textbf{Layers} & \textbf{Output}    \\ \hline
\textbf{Encoder}             & RGB Image (224x126) & \begin{tabular}[c]{@{}l@{}}inp, out, size, stride, padding\\ Conv2D(3, 64, 7, 2, 3), BatchNorm, ReLU\\ Resnet18 Layer1\\ Resnet18 Layer2\\ Conv2D(128, 128, 3, 2, 1), BatchNorm, ReLU\\ Conv2D(28, 256, 3, 2, 1), BatchNorm, ReLU\end{tabular} & Hidden State       \\ \hline
\textbf{Decoder}             & Hidden State        & \begin{tabular}[c]{@{}l@{}}inp, out, size, stride, padding\\ Deconv2D(256, 64, 4, 2, 1)\\ Deconv2D(64, 32, 4, 2, 1)\\ Deconv2D(32, 16, 4, 2, 1)\\ Deconv2D(16, 8, 4, 2, 1)\\ Deconv2D(8, 2, 4, 2, (2, 1))\end{tabular}                         & EgoHandMask \\ \hline
\textbf{Embedding Generator} & Hidden State        & \begin{tabular}[c]{@{}l@{}}inp, out\\ FullyConnected(7168, 2048), BatchNorm, ReLU\\ FullyConnected(2048, 83)\end{tabular}                                                                                                                      & Embedding          \\ \hline
\textbf{LSTM}                & Embedding           & \begin{tabular}[c]{@{}l@{}}inp, hidden, layers\\ LSTM(83, 83, 4)\\ FullyConnected(83, 83)\end{tabular}                                                                                                                                         & Class Label        \\ \hline
\end{tabular}
\end{center}
\caption{All the hyperparameters used in various components of our network.}
\label{tab:4_hyperparameters}
\end{table*}
\label{sec:4_network}
LSTMs have been widely used for video classification \cite{Wu2015}, action recognition \cite{Liu2017, zhu2016} and gesture recognition \cite{Molchanov2016, Cao2017}. In all the approaches above, the embeddings provided to LSTMs as input play a vital role in determining the recognition accuracy. In our architecture, embeddings based on segmented ego hands as described in Section \ref{sec:3_decon} provide inputs specific for ego hands to LSTMs, leading to better recognition accuracy. 

\subsection{Architecture}
\label{ssec:4_1_Architecture}
Our network architecture consists of two main components (Figure \ref{fig:3_network}), an autoencoder like network that generates segmented ego hand images and embeddings for LSTMs, and LSTMs that feed on this component to recognise the corresponding ego gesture. An autoencoder is a neural network that is intended to reproduce the input. Internally it reduces the input to a hidden state whose dimensions are lower than that of the input. Using this lower-dimensional hidden state, it tries to replicate the input. However, in our architecture, as described Section \ref{sec:3_decon}, instead of replicating the input, we generate a segmented ego hand image with the same dimensions as the input. This output maps each pixel in the input image to either ego-hand or context. In addition, we also output the embedding generated from the hidden state, whose dimensions are much smaller than the input image. 

For the encoder, we used the first two layers in ResNet18 \cite{Wu2016}, and further, we add 3 more convolutional layers, progressively decreasing the size of feature maps by setting the stride to 2 and simultaneously increasing the number of features to 256. Each of the convolutional layers is followed by batch normalisation and ReLU layers. We fix the input RGB image size to \(224*126*3\), which when passed through the encoder produces a hidden state of size \(7*4*256\). The decoder has a series of 5 deconvolutional filters, takes in the hidden state as input, upsamples the features back to the size of the image and simultaneously decreases the number of features. The \(7*4*256\) sized hidden state vector when passed through the decoder outputs a \(224*126*2\) dimensional image. Each of the two channels in the output contains the probability of the pixel belonging to ego-hand and the context. The complete set of parameters for the network is provided in the appendix for full reproducibility.

A sequence of 2 fully connected layers that reduce the dimensions of the flattened hidden state to the size of the LSTM input is added in a separate branch. The output of this branch is then fed as an input to LSTM layers. Empirically 4 LSTM layers with the same size for input and hidden state were found to yield the best result. The final hidden state from the fourth LSTM layer is connected to a fully connected layer to generate class probabilities for each gesture.

\begin{table*}[ht]
\begin{center}
\begin{tabular}{|l|c|c|c|}
\hline
Method & Modality & Frames & Accuracy \\
\hline\hline
\color{bad}
VGG16 + LSTM & RGB & Any & 0.747 \\
\color{bad}
VGG16 + LSTM & RGB & 40 & 0.808 \\
\color{bad}
VGG16 + RSTM (H) + LSTM  & RGB & 40 & 0.838 \\
\color{bad}
C3D + RSTTM (H) + LSTM & RGB & 40 & 0.893 \\
\color{bad}
C3D + RSTTM (H) + LSTM & Depth & 40 & 0.906 \\
\color{bad}
C3D + RSTTM (H) + LSTM & RGB + Depth & 40 & 0.922 \\
\hline
\color{good}
Segmentation based Embedding + LSTM (ours) & RGB & Any & \textbf{0.969} \\
\hline
\end{tabular}
\end{center}
\caption{Comparison of results on Ego Gesture dataset to the state of the art. The results reported from Cao \etal \cite{Cao2017} are in \textcolor{bad}{purple}. Our network (results reported in \textcolor{good}{green}) produces considerably better accuracy on just RGB data during inference and can use all the images in a sequence, while the state of the art is limited to 40 frames per sequence and needs both RGB and Depth data for best results.}
\label{tab:1_main_results}
\end{table*}

\subsection{Training Procedure}
\label{ssec:4_2_Training}
The network training is performed in 3 distinct steps. The first step consists of training the encoder, decoder and embedding generator together to output the segmented ego hand image and the input to LSTMs. As described in Section \ref{sec:3_decon}, we created segmented ego hand images from depth images to avoid manual annotation for ground truth data. In this training step, each image is considered an individual entity. We shuffle all the images with ego hands without respecting their order in a video, split the total images into training, validation and testing sets. We use two loss functions, one to control the generation of segmented ego hand images and another to label each embedding with the corresponding gesture. Cross-entropy loss for segmented ego hand images and label loss with equal weights are used for backpropagation. 

The LSTM layers are trained in the second step to recognise ego hand gestures from sequences of embeddings. The videos each containing a single gesture are split into training, validation and testing sets. Embeddings for videos are generated using the network weights from step 1. These embeddings are then used to train the LSTMs. Isolating LSTM training allowed us to use bigger batch sizes which helped in better generalisation and also to use sequences of arbitrary length. Cross entropy loss is used on the final fully connected layer to classify each sequence.

In the final step, we train the entire network together end-to-end by connecting the embedding generator branch to LSTMs as input. We initialise the encoder, decoder, embedding generator with weights from step 1 and LSTMs with weights from step 2. The final loss function is the sum of Cross-Entropy label loss from LSTMs and segmented ego hand images loss from the decoder are. All the hyperparameters and initialisation strategies used are discussed in Section \ref{sec:5_exp}.

\begin{figure*}[htb!]
\begin{center}
\includegraphics[scale=0.8]{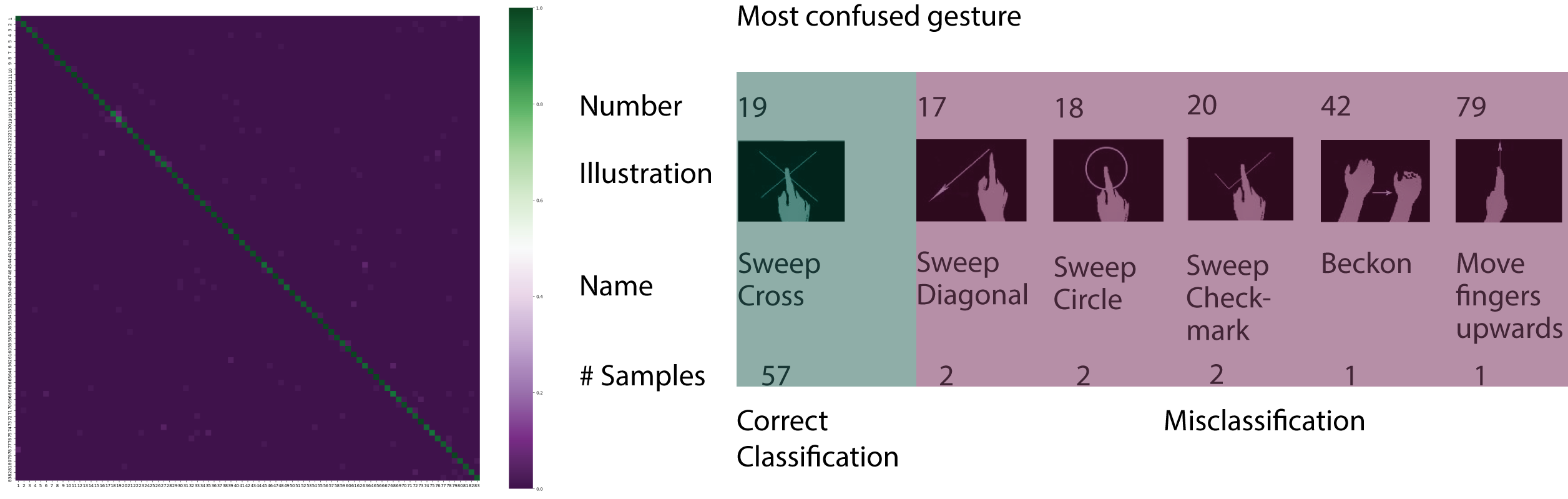}
\end{center}
   \caption{Confusion matrix for all the gestures. The most confused gesture as seen is Sweep Cross(19), out of the 65 samples 57 were correctly classified while 2 each were misclassified as Sweep Diagonal, Sweep Circle,  Sweep Check-mark and 1 each as Beckon and Move Fingers Upwards.}
\label{fig:1_confusion_matrix}
\end{figure*}

\section{Experiments and Results}
\label{sec:5_exp}

The network architecture proposed in Section \ref{sec:4_network} was tested and validated on EgoGesture dataset. In the following sections  we describe the dataset, hardware setup used to run the validation and experiments, results compared to the state of the art and validation done through ablation studies.

\subsection{Dataset and Experimental Setup}
\label{ssec:5_1}
The EgoGesture dataset \cite{Zhang2018} is the only publicly available dataset that has a large number of videos recorded with first-person view cameras performing ego hand gestures. 50 subjects performed 83 different hand gestures in both indoor and outdoor environments. The dataset contains more than 24,000 video samples which are collected in 6 different environments having a large variation in illumination and duration per gesture. It also captures both RGB and depth modalities. These properties of the EgoGesture dataset makes it a valid candidate for testing and bench-marking ego gesture recognition algorithms. Though there is another dataset made specifically for ego gestures \cite{tejo2018}, it only contains 10 gestures. Since the EgoGesture dataset has more variety and variations, we focused our experiments on this dataset. 

We used PyTorch for software, NVIDIA Titan Xp and NVIDIA RTX 2080 Ti for hardware to create and train the networks. The training procedure followed is detailed in Section \ref{ssec:4_2_Training}. In the first step, we trained the encoder, decoder and embedding generator. The images with ego hand are divided into training, validation, testing sets with 0.6, 0.2, 0.2 splits yielding \(536938, 178979, 178978\) images respectively. The ResNet18 layer 1, layer 2 are initialised with pre-trained ImageNet weights, the rest of the weights and biases are initialised with zeros. We used a batch size of \(100\), set the learning rate to \(1e-6\) and used ADAM optimiser. Step 1 of the training is done until the validation accuracy does not improve. The test set accuracy was used as an intermediary validation step. Figure \ref{fig:2_Decon} shows some ego hand segmentations generated by our decoder after the first step of training.
\begin{table}
\begin{center}
\begin{tabular}{|l|c|c|}
\hline
Scenario & State of the Art \cite{Cao2017} & Our Network \\
\hline\hline
Walking & 0.828 & 0.962 \\
Stationary & 0.866 & 0.972 \\
\hline
\end{tabular}
\end{center}
\caption{Accuracy on scenarios with (walking) and without (stationary) ego-motion. We outperform the state of the art in both scenarios.}
\label{tab:2_ego_motion_analysis}
\end{table}

We experimented with different initialisation techniques for LSTMs in the second stage of training and found a combination of orthogonal \cite{saxe2013} and Xavier normal \cite{glorot2010} initialisation for input weights and recurrent weights, zeros for biases worked best for convergence. The embeddings created are stored on disk and used as input to this step. Unlike the state of the art \cite{Cao2017}, whose network is limited to identifying gestures from a sequence of length 40, this step gave us the ability to train with arbitrary sequence length and large batch size. We used a batch size of 100 with padded sequences for training the LSTM layers. We set the input size and hidden size for LSTM layers to \(83\). The segmented ego hand gesture videos are divided into training, validation, testing sets with 0.6, 0.2, 0.2 splits yielding \(14495, 4831, 4831\) samples respectively. We trained in a step-wise manner using \(1e-2\) learning rate until the training loss started to diverge. At this step, we further decreased the learning rate to \(1e-3\) and trained until validation accuracy did not improve anymore. The network and the trained weights are available on \href{https://github.com/V-Sense/ssar}{github}.

In the final step, we combined the encoder, decoder, embedding generator and the LSTMs layers and trained with one image sequence per batch. Weights from the first and second steps are used to initialise the network. The entire network is trained with label loss from the LSTM layers and segmentation loss from the decoder. Since the two losses were trained individually before this step, using the sum of two losses was adequate for the final step. We used ADAM optimiser with a learning rate of \(1e-3\) and trained until the validation accuracy did not improve and reported test accuracy.

\subsection{Results}
\label{ssec:5_2}
The right embeddings to LSTM layers influence their recognition capacity to a large extent. This is empirically evident from Table 
\ref{tab:1_main_results}, where the differentiating factors that influence the accuracy are the embedding inputs to LSTMs. Cao \etal \cite{Cao2017} argue adding homographic spatial transformer modules to VGG embeddings and homographic recurrent spatio-temporal transformer modules to C3D embeddings result in better embeddings to LSTMs, increasing their recognition accuracy. However, we demonstrate that by using segmentation based embeddings we can get considerably better accuracy for the EgoGesture dataset. We posit that this is possible since any feature that does not belong to ego hands is ignored by design in our network. Embeddings created by our network are visualised in Figure \ref{fig:4_visualisation}, as it can be seen that features around ego hands contributed most for the Gradient-Class Activation Map (Grad-CAM images)\cite{SelvarajuDVCPB16}. For gestures like Number6 fingers on ego hands become important for recognition, our network as seen in Figure \ref{fig:4_visualisation} correctly paid most attention to fingers in the RBG images. In case of two handed gestures, the activation maps show that attention is being paid to both the hands as it would be expected for recognising the gesture.

The most confused gesture in our analysis using a confusion matrix (Figure \ref{fig:1_confusion_matrix}) was the Sweep Cross. Out of the 65 test sample for this gesture, 57 were correctly classified, while 2 each were misclassified as Sweep Diagonal, Sweep Circle and Sweep Check-mark and 1 each as Beckon and Move Fingers upward. Since these gestures look quite similar, it is very probable for them to confused with each other.

We grouped the gestures into two sets, one containing ego-motion(walking) and another containing no ego-motion (stationary), following the procedure from the state of the art. The results reported in Table \ref{tab:2_ego_motion_analysis} show that using segmentation based embeddings is sufficient to compensate for ego-motion. Our network performed better in both stationary and walking scenarios. The difference in accuracy between stationary and walking scenarios in our case is \(1\%\), whereas it is \(3.8\%\) in the case of the state of the art, which further illustrates that our network by not paying attention to the context around ego hands can deal with ego-motion better.

\subsection{Validation}
\label{ssec:5_3}\textbf{}
To validate the usage of segmentation based embeddings, we did ablation studies and compared them to the results reported by Cao \etal \cite{Cao2017}. We used the encoder part of the network, connected it to the embedding generator and then LSTM, we did not use the decoder to make sure that there is no influence from attempting to recover ego hand masks. We trained this network end-to-end to recognise ego gesture images with cross-entropy loss. The back-propagation was performed similarly to other training until validation accuracy didn't improve. Our recognition accuracy from this simple embeddings + LSTM network (0.754) was very close to VGG16+LSTM (0.747) as reported in Table \ref{tab:3_ablation_results}. To further understand the effect of using segmentation based embeddings we tested and reported the results of recognition accuracy on our network after performing phase one and two of training as described in Section \ref{ssec:4_2_Training}. After training the encoder and decoder, we created segmentation based embeddings, which were used as inputs for LSTM training. We skipped the final fine-tuning step mentioned Section \ref{ssec:4_2_Training}. The recognition accuracy improved considerably compared to the above two networks thus validating the usage of segmentation based embeddings. We posit that these embeddings carry information particular to ego hands, which lead to better generalisation capability than compared to simple embeddings as evident with the improved accuracy. The accuracy of the three networks are reported in Table \ref{tab:3_ablation_results}. When we performed the final stage of training, the decoder performing ego hand mask generation acts as a regulariser further improving the performance and giving us the best result.

\begin{table}
\begin{center}
\begin{tabular}{|l|c|}
\hline
Method & Accuracy \\
\hline\hline
VGG16 + LSTM\cite{Cao2017} & 0.747 \\
Simple Embedding + LSTM & 0.754   \\
Segmentation based Embedding + LSTM & 0.947 \\
\hline
\end{tabular}
\end{center}
\caption{Table validating the use of segmentation based embedding. We use the encoder as described in section \ref{ssec:4_1_Architecture} to generate simple embeddings for the LSTM to recognise an ego gesture. In comparison to VGG16 + LSTM, this does not result in a significantly increased accuracy. The segmentation based embedding + LSTM approach, performs only the first part and second part of the training described in section \ref{ssec:4_2_Training}. We forgo the final step of training to isolate the effect of using segmentation based embedding. The accuracy increases, when compared to the simple embeddings, which validates the use of segmentation based embeddings.}
\label{tab:3_ablation_results}
\end{table}

\section{Conclusions and Future Work}
\label{sec:6_concl}

We propose the concept of segmentation based embeddings for ego hands and a deep neural network architecture that creates and uses them to recognise ego gestures. We used a three-step training process which facilitates training our network architecture with large batch sizes. Our results demonstrate the proposed architecture can deal with ego-motion and recognise ego gestures considerably better than the state of the art while using only RGB modality during inference.

In the future, we would like to explore adopting the concept of segmentation based embeddings to other complex ego vision tasks like Ego Action recognition using deep learning and measure our performance on datasets like EGTEA Gaze+ \cite{li2018}, Something Something dataset \cite{goyal2017} and Epic Kitchens \cite{damen2018}.

\section*{Acknowledgements}
This publication has emanated from research conducted with the financial support of Science Foundation Ireland (SFI) under the Grant Number 15/RP/2776.

\end{document}